# From Cheap to Pro: A Learning-based Adaptive Camera Parameter Network for Professional-style Imaging

Fuchen Li[1,#], Yansong Du[2,#], Wenbo Cheng[1], Xiaoxia Zhou[1], Sen Yin[3,4,*]

*Abstract*— Consumer-grade camera systems often struggle to maintain stable imaging quality under complex illumination conditions such as low-light, high dynamic range, backlighting, and spatial color temperature variations. These challenges lead to underexposure, color casts, and tonal inconsistencies, which degrade the performance of downstream vision tasks. Traditional rule-based auto-parameter modules lack scene awareness and adaptability, showing limited effectiveness in dynamic environments. To address this, we propose ACamera-Net, a lightweight and scene-adaptive camera parameter adjustment network that directly predicts optimal exposure and white balance from RAW inputs. The framework consists of two modules: ACamera-Exposure, which estimates ISO to alleviate underexposure and contrast loss, and ACamera-Color, which predicts correlated color temperature and gain factors for improved color consistency. Optimized for real-time inference on edge devices, ACamera-Net can be seamlessly integrated into imaging pipelines. Trained on diverse real-world data with annotated references, the model generalizes well across lighting conditions. Extensive experiments demonstrate that ACamera-Net consistently enhances image quality and stabilizes perception outputs, outperforming conventional auto-modes and lightweight baselines without relying on additional image enhancement modules.

## I. INTRODUCTION

Visual perception is a core function of modern camera systems, supporting critical tasks such as navigation [1], localization [2]–[4], and object detection [5]–[7]. However, in real-world scenarios, complex lighting conditions—such as low illumination, high dynamic range scenes, backlighting, and spatially varying color temperatures—can severely degrade image quality, leading to artifacts such as underexposure, color casts, and tonal inconsistency [8]. These issues negatively affect the performance of downstream perception modules. Although most commercial cameras are equipped with automatic exposure and white balance functions, these modules are typically rule-based [9], lacking scene awareness and adaptability, and often perform poorly under varying environments, especially in real-time or embedded applications [10]. While some high-end devices provide more advanced auto-adjustment capabilities, their high cost limits widespread adoption [11]. Consequently, many perception systems still rely on suboptimal visual input, resulting in unstable detection, reduced localization accuracy, and inconsistent control behaviors, ultimately compromising overall system reliability.

[#]These authors contributed equally to this work and are considered co–first authors.
[*]Corresponding author: senyin@polyu.edu.hk
[1]*Xingxi Technology Co., Ltd* [2]*Tsinghua University* [3]*The Hong Kong Polytechnic University* [4]*Henan University of Science and Technology*

We propose ACamera-Net, a lightweight and scene-adaptive camera parameter adjustment network that directly optimizes visual sensing quality during image capture. Unlike traditional approaches that rely on fixed settings or post-capture enhancement, ACamera-Net predicts optimal exposure and white balance parameters from RAW inputs in real time, producing well-exposed and color-balanced images without the need for extensive post-processing. The system consists of two functional modules: ACamera-Exposure, which adaptively estimates the optimal ISO to improve brightness and contrast, and ACamera-Color, which predicts white balance parameters—including correlated color temperature and red/blue channel gains—by modeling global and local chromatic features. Designed for efficiency and deployability, the framework is well-suited for edge inference on resource-constrained platforms.

We integrate the proposed framework into a mobile camera system and validate its effectiveness in static image capture tasks. Experimental results show that our method significantly improves image quality and visual consistency across various environments, demonstrating strong practicality and robustness. The main contributions of this work are as follows: we propose ACamera-Net, a real-time, adaptive camera parameter adjustment framework capable of generating high-quality, professional-style images under diverse lighting conditions without manual intervention; we design two functional modules—ACamera-Exposure and ACamera-Color—which respectively predict optimal ISO and white balance parameters based on luminance and chromatic features extracted from RAW inputs; and we deploy our method on real-world camera platforms, achieving consistent improvements over existing auto-parameter baselines and lightweight learning-based methods in terms of image quality and visual consistency.

## II. RELATED WORK

Robust visual perception under diverse and often challenging lighting conditions is essential not only for professional applications such as navigation, manipulation, and human-computer interaction [12], [13], but also for everyday consumer photography and videography [14]. Users frequently expect their cameras to deliver clear, natural, and aesthetically pleasing results across scenarios ranging from dim indoor settings to high-contrast outdoor scenes [15]. However, achieving consistently high-quality visual input in such dynamic and uncontrolled environments remains a significant challenge, especially for resource-constrained consumer devices [16]. This section reviews recent advances in exposure and color adjustment techniques, emphasizing their limitations when applied to practical camera applications.

## A. Traditional Automatic Camera Adjustment

Most consumer and professional cameras rely on built-in automatic exposure (AE) and automatic white balance (AWB) modules to regulate image quality [17]. These modules are typically based on fixed-rule heuristics, which perform well under standard lighting conditions but often fail in complex scenarios involving mixed illumination, backlighting, or abrupt brightness transitions. Due to their limited scene understanding and lack of contextual adaptation, such systems frequently produce suboptimal visual outputs, negatively impacting downstream image processing and computer vision tasks.

High-end imaging systems—such as full-frame and medium-format cameras—address these challenges through a range of hardware-level enhancements, including real-time histogram analysis, exposure bracketing, high-bit-depth RAW processing, and proprietary color science [18]. However, these features are rarely available or practical in most camera platforms, which are typically constrained by cost, power consumption, and computational resources. Moreover, general users often lack the expertise to manually adjust parameters in response to varying scene conditions [19].

## B. Learning-Based Image Enhancement and ISP Replacement

Early research introduced learning-based alternatives to overcome the limitations of traditional image signal processing (ISP) pipelines. For instance, convolutional neural networks (CNNs) have been applied to tasks such as demosaicing, denoising, and exposure correction [20]. While these approaches improve perceptual quality in static image enhancement scenarios, they are typically applied post-capture and offer limited support for real-time control or integration into practical camera systems. Moreover, most of these models are designed as isolated functional replacements, without considering the global dependencies among camera parameters [21].

Subsequent studies explored generative adversarial network (GAN)-based frameworks for aesthetic enhancement and image style transfer. Methods such as DualAST [22] and Structure-Guided Transfer [23] improve style consistency and perceptual quality, but remain limited to offline post-processing workflows. As a result, they are not well-suited for deployment in real-world camera systems where adaptive and on-device control is required.

## C. Deep Parameter Prediction and Feedback-Based Control

More recent approaches have explored scene-aware parameter prediction, aiming to learn exposure and color settings directly from visual context. For example, scene-classification-based models have been applied to exposure control [24], and several works jointly predict denoising strength or color matrices [25]. Other methods adopt feedback-driven ISP configuration strategies, such as AdaptiveISP [26] and DynamicISP [27], which adjust image processing parameters through learned feedback loops. While these approaches improve adaptability, they often rely on highly complex black-box architectures, making them difficult to interpret and deploy on practical camera platforms.

In parallel, learning-based control has also been extended to traditional 3A modules. For instance, PhotoHelper provides real-time composition suggestions using deep feature retrieval [28], and AQA models [29] incorporate photographic principles into quality prediction. However, most of these systems focus on narrow, isolated tasks and lack a unified framework for scene-adaptive, style-consistent global parameter control.

## D. Summary

Despite notable progress in perception-aware tuning and data-driven enhancement, existing solutions often face three major limitations when applied to practical camera systems: (1) a lack of end-to-end integration within the image acquisition pipeline, (2) high computational complexity and limited interpretability, and (3) insufficient control over global visual style, particularly under dynamic lighting conditions. To address these challenges, we propose a lightweight, interpretable, and deployable solution that enables real-time prediction of camera parameters during image capture, allowing cameras to consistently produce high-quality visual outputs across diverse environments.

## III. ACAMERA NETWORK ARCHITECTURE DESIGN

This section presents the architectural design of the proposed ACamera network, aiming to create a lightweight, deployable system for automatic camera parameter adjustment. The system dynamically adjusts key imaging parameters—ISO, white balance temperature, and channel-specific gains—according to lighting conditions and stylistic requirements, enhancing brightness structure, color consistency, and visual aesthetics. Unlike traditional photography that requires expert experience and advanced hardware, our method operates in real time on consumer-level platforms, such as mobile devices and robotics systems.

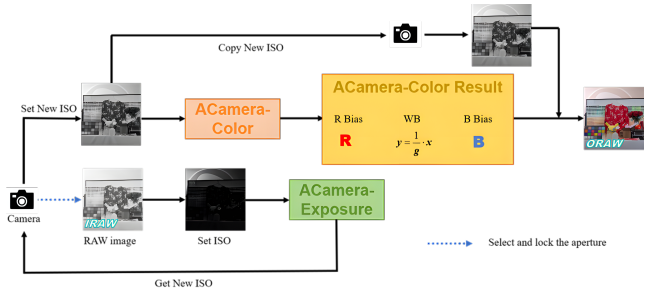

Fig. 1: Overview of the proposed ACamera framework. It mainly consists of two modules: ACamera-Exposure and ACamera-Color, with RAW as input for easy deployment in various camera pipelines.

ACamera consists of two core sub-modules: ACamera-Exposure and ACamera-Color, responsible for exposure control and color modulation respectively. The system takes an initial RAW image as input and passes it through the exposure module to predict the optimal ISO. A second image is then captured using the predicted ISO and sent to the color module for white balance and gain estimation. A parameter-aware modulation module coordinates both sub-networks to further enhance adaptability and style control. The full pipeline is illustrated in Fig. 1.

### A. ACamera-Exposure Module

The ACamera-Exposure module predicts the optimal ISO setting based on an initial RAW image captured with a fixed ISO (set to 1000). It processes the input through several convolutional layers followed by residual blocks to extract brightness-aware features, which are then mapped into a scalar ISO prediction. The module architecture is depicted in Fig. 2.

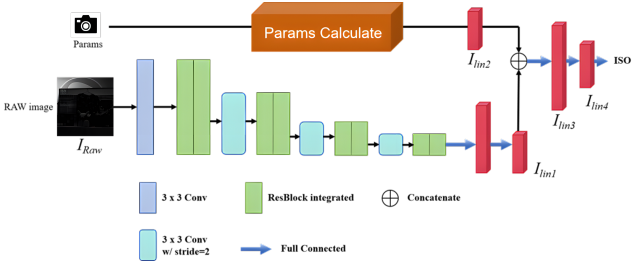

Fig. 2: Overview of the proposed ACamera-Exposure framework. Params Calculate module is purpose to convert the initialized optical parameters into linear layers.

Given that ISO is a continuous but discretely operated parameter, we reformulate the regression task as a distribution learning problem using the *Distribution-Enhanced Loss (DEL)*. Specifically, we discretize the ISO range into bins $\{b_1, b_2, ..., b_n\}$ and let the model predict a soft distribution $\mathbf{P} = (p_1, ..., p_n)$ over them. The final ISO is estimated by:

$$y = \sum_{i=1}^{n} p_i \cdot b_i \quad (1)$$

To emphasize bins near the ground truth ISO during training, we define a soft-weighted cross-entropy:

$$L_{\text{DEL}} = -\sum_{i=1}^{n} w_i \log(p_i), \quad w_i = \max\left(0.1, 1 - \frac{|y - b_i|}{\Delta}\right) \quad (2)$$

This approach stabilizes training and improves local brightness prediction accuracy.

### B. ACamera-Color Module

The ACamera-Color module takes a second RAW image, captured with the predicted ISO, and estimates three color-related parameters: color temperature (Temp), red channel bias ($\Delta R$), and blue channel bias ($\Delta B$). The image is first decomposed into four CFA channels (R, Gr, Gb, B) and processed via separate convolutional and residual blocks. The resulting features are pooled and concatenated to form a unified representation (Fig. 3). The module outputs are

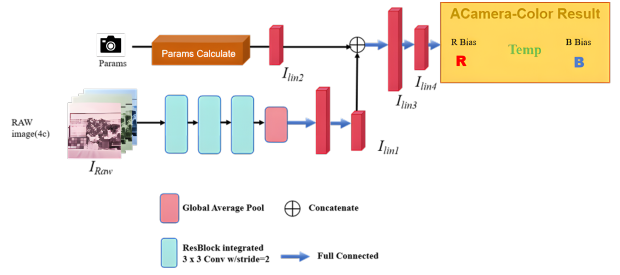

Fig. 3: Overview of the proposed ACamera-Color framework.

converted to final gain values using:

$$R_{\text{gain}} = \frac{R_{\text{ref}} + \Delta R}{R_{\text{measured}}}, \quad B_{\text{gain}} = \frac{B_{\text{ref}} + \Delta B}{B_{\text{measured}}} \quad (3)$$

Here, $R_{\text{ref}}$ and $B_{\text{ref}}$ are reference values while $R_{\text{measured}}$ and $B_{\text{measured}}$ are average channel values. This design maintains color fidelity while providing fine control over stylistic tonality.

### C. Parameter-Aware Modulation

To model the influence of physical capture conditions—such as shutter speed, aperture, and focal length—we introduce a parameter-aware modulation mechanism. This module encodes imaging parameters as auxiliary features and injects them into the main network branches to guide feature processing (Fig. 4.). Given an imaging parameter $a$,

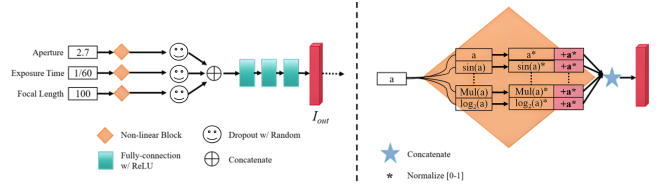

Fig. 4: Overview of the proposed ACamera-Color framework.

we normalize it to a bounded scalar $a^* \in [0, 1]$ and apply a residual fusion:

$$\tilde{a} = a^* + a \quad (4)$$

Here, $a$ denotes the raw value (e.g., shutter time in ms), and $a^*$ is its normalized counterpart. The fused representation $\tilde{a}$ preserves the physical semantics of $a$ while ensuring numerical stability during learning. This design enhances the network's robustness under varying capture settings.

To avoid overfitting to specific input parameters, we apply random channel drop during training, randomly masking part of the parameter set.

### D. Joint Training and Loss Design

The ACamera network is trained end-to-end using a multi-branch supervision scheme. Each predicted parameter—ISO, color temperature, and color gain—has a corresponding loss term. The total loss is defined as:

$$\mathcal{L}_{\text{total}} = \lambda_1 \cdot \mathcal{L}_{\text{exp}} + \lambda_2 \cdot \mathcal{L}_{\text{color}} + \lambda_3 \cdot \mathcal{L}_{\text{mod}} \quad (5)$$

where $\lambda_1$, $\lambda_2$, and $\lambda_3$ are the respective loss weights, initially set equally and adjusted dynamically based on perceptual error during training.

The network is optimized using Adam with an initial learning rate of $1 \times 10^{-4}$, decayed over 100 epochs. Training is conducted on a RAW dataset with ISO and white balance ground truth. We use a two-stage strategy: pretraining the Exposure module followed by joint training with the Color and Modulation branches.

Our final model contains fewer than 2.3M parameters and achieves inference latency under 20ms on embedded platforms such as Jetson Orin and Snapdragon SoCs, demonstrating real-time performance and deployment feasibility.

## IV. EXPERIMENTS AND PERFORMANCE EVALUATION

### A. Data Collection and Processing

To train and evaluate our ACamera-Net framework under diverse real-world conditions, we curated a multi-scene dataset containing synchronized RAW and RGB image pairs, as shown in Fig. 5. The data was collected using a low-cost imaging system (OmniVision OV4689 sensor) mounted on a mobile robotic platform across representative indoor and outdoor scenes, including dim corridors, overexposed streets, mixed illumination offices, and HDR environments. All frames were stored in RAW format to retain the full signal range for accurate exposure and white balance learning. To ensure robust learning and consistent evaluation, we implemented a data construction pipeline consisting of three stages: sampling, annotation, and quantization. In the sampling stage, we uniformly selected representative frames from continuous video streams. During annotation, professional photographers labeled the ground-truth ISO values and white balance parameters (Correlated Color Temperature and red/blue channel gains) to reflect optimal perceptual quality. We then discretized these continuous labels into categorical classes using a soft interval encoding strategy, enhancing model convergence and prediction stability.

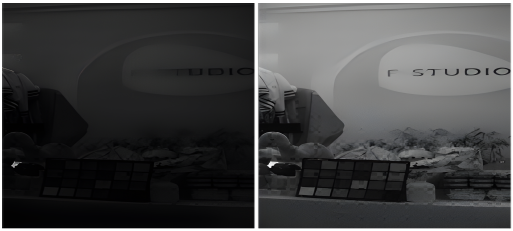

Fig. 5: The data is acquired with ISO transformations and the results are visualized on RAW .

In total, our dataset contains over 12,000 labeled samples across more than 25 scenes, including both synthetic and real-world conditions. This diverse and well-annotated dataset provides a solid foundation for training scene-adaptive camera parameter regression models.

### B. Implementation Details

We trained our model using PyTorch on an NVIDIA RTX 4090 GPU with mixed precision. The network was optimized with the AdamW optimizer (initial LR=1e-4, batch size=8) and trained end-to-end on our curated dataset of RAW Bayer images.

To validate real-world deployment, the trained model was quantized to INT8 and ported to a HiSilicon Hi3516DV300 SoC, which features integrated NPU acceleration and embedded ISP support. We used the HiAI DDK and ACL SDK to convert and optimize the model for edge inference. The architecture of our proposed system is illustrated in Fig. **??**. For evaluation, we built a low-cost imaging system using a 1/4" CMOS sensor (OmniVision OV4689), commonly found in budget surveillance or automotive dashcam modules. This sensor outputs RAW10 Bayer frames at 1080p@30fps via MIPI-CSI. Compared to high-end sensors, it suffers from considerable fixed-pattern noise, lower dynamic range, and degraded color fidelity—particularly under challenging lighting conditions. These limitations make it ideal for demonstrating the effectiveness of our method.

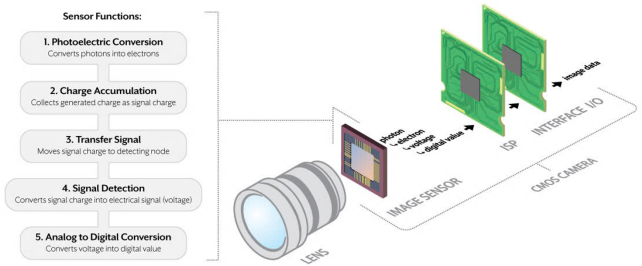

Fig. 6: Architecture of the Custom AI Camera System .

### C. Evaluation of Exposure Prediction Module

To evaluate the effectiveness of the proposed ACamera-Exposure module under real-world conditions, we conduct extensive experiments across three challenging scenarios: low-light, high dynamic range (HDR), and backlighting. As illustrated in Fig. 7, we compare four types of results: (1) images captured using the default automatic mode on standard consumer devices (Auto Normal); (2) results obtained from professional cameras in auto mode (Auto Pro); (3) outputs generated by our proposed ACamera-Exposure module (Ours); and (4) manually tuned reference images (GT).

Qualitatively, Auto Normal often suffers from underexposure in dark scenes and highlight clipping in HDR environments. Auto Pro shows partial improvements but still struggles to deliver consistently optimal results. In contrast, our method effectively preserves both shadow and highlight details, achieving a visually balanced appearance closer to the professional ground truth.

To complement the visual comparison, we introduce objective metrics, as summarized in Fig. 8. Specifically, we compare the estimated ISO values across methods to assess the exposure prediction accuracy. In low-light and backlighting scenarios, our method significantly improves ISO estimation, reducing the discrepancy with ground truth. For instance, under low-light, the predicted ISO using Auto Normal is

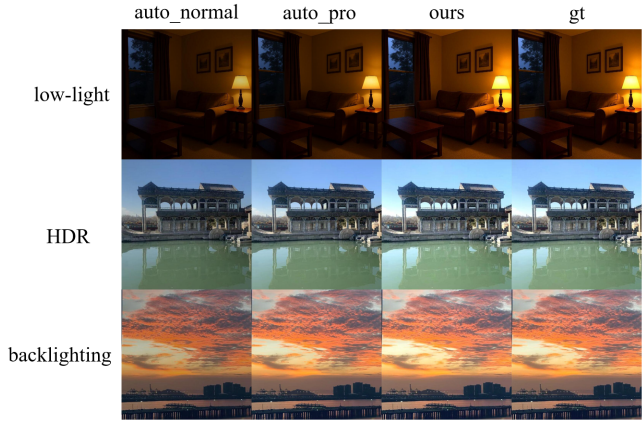

Fig. 7: ACamera-Exposure results and Other Methods under challenging lighting conditions.

400, while our method reaches 750, aligning closely with the GT value of 750.

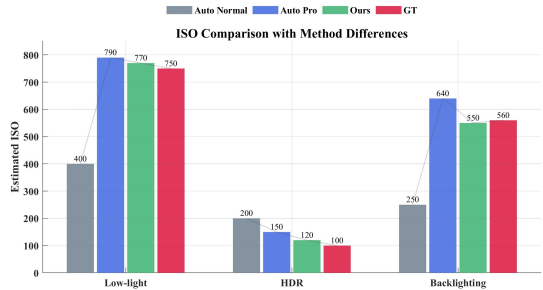

Fig. 8: ISO testing results across different scenes under various methods.

In addition, we measure the Luminance Deviation across different lighting conditions to evaluate brightness consistency. As shown in Fig. 9, our approach achieves the lowest luminance deviation across all three settings (e.g., 3.6 in backlighting and 3.8 in low-light), indicating superior exposure regularity and visual comfort.

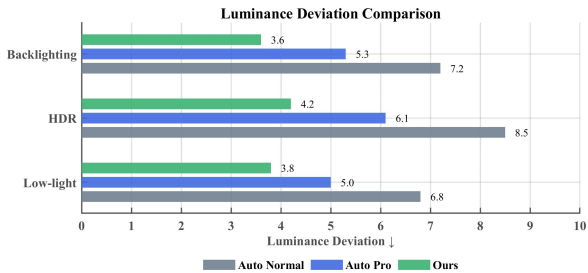

Fig. 9: Luminance Deviation results across different scenes under various methods.

### D. Evaluation of White Balance Prediction Module

We validate the effectiveness of the proposed ACamera-Color adjustment module under three representative lighting conditions: daylight, warm light, and cool light. As illustrated in Fig. 10, the proposed method delivers more faithful and visually consistent color reproduction compared to standard baselines, including (1) default auto white balance from a consumer camera (auto_normal) and (2) enhanced auto mode from a flagship smartphone (auto_pro). Noticeable color distortions—such as yellow shifts under daylight and blue casts under cool light—persist in auto_normal outputs, while auto_pro achieves partial correction but still struggles in complex scenarios. In contrast, the proposed method generates neutral whites and accurate tonal balance, closely resembling the manually adjusted ground truth.

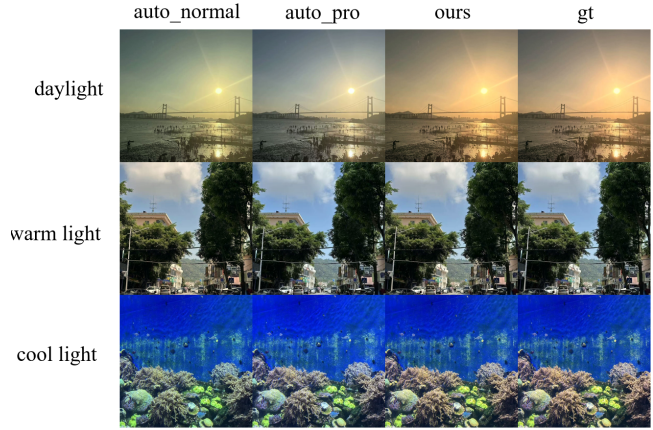

Fig. 10: ACamera-Color results and Other Methods under challenging lighting conditions.

Quantitative evaluation is conducted using the average CIE $\Delta E$ metric to assess perceptual color difference against the GT. As shown in Fig. 11, the proposed approach achieves the lowest $\Delta E$ scores across all lighting conditions. Under daylight and warm light, auto_normal yields high errors of 18 and 23 respectively, while auto_pro reduces these to 12 and 15. However, both remain above the perceptual threshold. The proposed method maintains $\Delta E$ values below 5 in all settings, indicating its superior robustness and color accuracy even under challenging illumination.

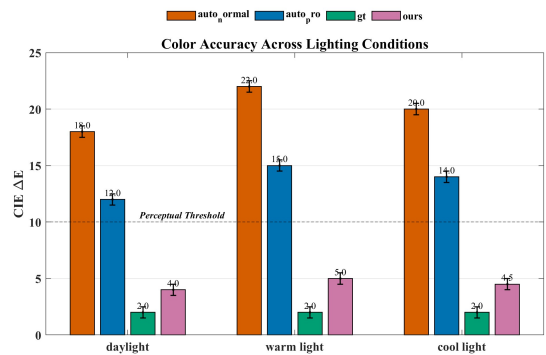

Fig. 11: CIE $\Delta E$ results across different scenes under various methods.

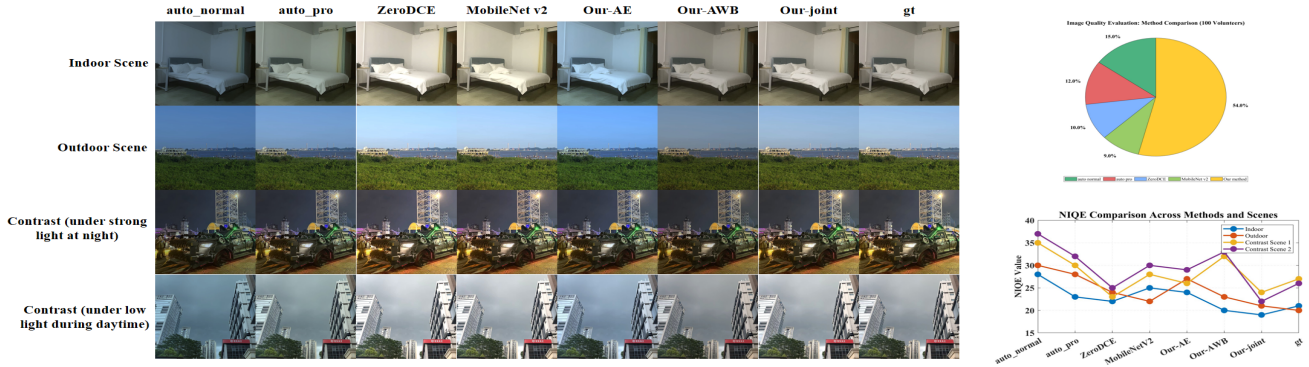

Fig. 12: Final results of our method and comparative analysis.

### E. Evaluation of Joint ACamera-Net

We combined the ACamera-Exposure and ACamera-Color modules and conducted a series of comparison experiments to verify the effect of the ACamera-Net network on improving image quality. In addition to comparing with the original sensor images (auto_normal) and high-end sensor images (auto_pro), we also compared the joint model with individual adaptive exposure module (ACamera-Exposure), adaptive white balance module (ACamera-Color), and existing image enhancement methods such as ZeroDCE [30] and MobileNet v2 [31]. The experiments covered four typical scenes: indoor scene, outdoor scene, strong light at night, and low light during the day. To comprehensively evaluate the performance of these methods, we adopted both subjective and objective evaluation methods. In the subjective evaluation, we invited 100 volunteers to rate the images processed by different methods. The experimental results showed that 54% of the respondents thought our images were closer to the ground truth and more in line with natural human aesthetics, showing superior visual effects compared to other methods. In the objective evaluation, we used Natural Image Quality Evaluator (NIQE) as the main evaluation metric, where lower NIQE values indicate better image quality. After calculation, the NIQE values for our method in the four scenes were: 19 (indoor scene), 21 (outdoor scene), 24 (strong light at night), and 22 (low light during the day). These results indicate that the error rate compared to the ground truth (gt) value is only 8.51%, and is lower than other methods as well as our individual modules. This demonstrates that our joint model performs excellently across multiple scenes, effectively improving image quality, especially under complex lighting conditions, providing more natural and realistic image results. Our result is as shown in Fig. 12.

## V. SUMMARY AND OUTLOOK

This paper presents ACamera-Net, an adaptive visual perception framework that directly operates on RAW images, specifically designed to enhance image quality and perceptual stability in consumer-grade camera systems under complex environmental conditions. The proposed framework consists of two lightweight modules responsible for ISO and white balance (color temperature) prediction, enabling end-to-end optimization of input images without requiring any hardware modifications. Extensive experiments across diverse representative scenes demonstrate that our method significantly outperforms traditional auto-adjustment systems and existing baseline networks in terms of luminance consistency, color fidelity, and structural preservation. Ablation studies further validate the individual contributions of each module, while the joint model exhibits superior robustness and generalization in adaptive imaging tasks. Overall, ACamera-Net provides an efficient and deployable solution for improving the visual input quality of consumer-grade cameras, laying a solid foundation for downstream tasks such as object detection, semantic understanding, and visual navigation. In future work, we plan to extend this framework to streaming video, integrate spatiotemporal modeling, and explore cross-modal adaptations with inertial or depth signals, thereby advancing the development of robust, intelligent imaging systems for dynamic real-world environments.

## ACKNOWLEDGMENT

The authors would like to thank Xingxi Technology Co., Ltd. for providing industrial support and deployment environment during the system development stage. This work was also partially supported by the Hong Kong Polytechnic University and Henan University of Science and Technology. Special thanks to Dr. Yin Sen for his guidance and coordination across multiple institutions.

## REFERENCES

[1] I. A. Kazerouni, L. Fitzgerald, G. Dooly, and D. Toal, "A survey of state-of-the-art on visual slam," *Expert Systems with Applications*, vol. 205, p. 117734, 2022.
[2] Y. Du, Z. Jiang, J. Tian, and X. Guan, "Modeling, analysis, and optimization of random error in indirect time-of-flight camera," *Optics Express*, vol. 33, no. 2, pp. 1983–1994, 2025.
[3] Y. Du, Y. Deng, Y. Zhou, F. Jiao, B. Wang, Z. Xu, Z. Jiang, and X. Guan, "Multipath interference suppression in indirect time-of-flight imaging via a novel compressed sensing framework," *arXiv preprint arXiv:2507.19546*, 2025.
[4] Y. Du, Y. Deng, Y. Zhou, F. Jiao, J. Song, and X. Guan, "Towards high-precision depth sensing via monocular-aided itof and rgb integration," *arXiv preprint arXiv:2508.16579*, 2025.
[5] Y. Du, J. Yao, F. Jiao, Y. Zhou, Q. Jin, B. Wang, K. An, Z. Jiang, and X. Guan, "A new method for removing internal scattering noise in itof camera," in *2025 Conference on Lasers and Electro-Optics Europe & European Quantum Electronics Conference (CLEO/Europe-EQEC)*. IEEE, 2025, pp. 1–1.

[6] M. Hussain, "Yolo-v1 to yolo-v8, the rise of yolo and its complementary nature toward digital manufacturing and industrial defect detection," *Machines*, vol. 11, no. 7, p. 677, 2023.

[7] J. Terven, D.-M. Córdova-Esparza, and J.-A. Romero-González, "A comprehensive review of yolo architectures in computer vision: From yolov1 to yolov8 and yolo-nas," *Machine learning and knowledge extraction*, vol. 5, no. 4, pp. 1680–1716, 2023.

[8] L. Agnolucci, L. Galteri, M. Bertini, and A. Del Bimbo, "Arniqa: Learning distortion manifold for image quality assessment," in *Proceedings of the IEEE/CVF Winter Conference on Applications of Computer Vision*, 2024, pp. 189–198.

[9] L. A. Swarowsky, R. F. Pereira, and L. B. Durand, "Influence of image file and white balance on photographic color assessment," *The Journal of Prosthetic Dentistry*, vol. 133, no. 3, pp. 847–856, 2025.

[10] F. Palermo, L. Casciano, L. Demagh, A. Teliti, N. Antonello, G. Gervasoni, H. H. Y. Shalby, M. B. Paracchini, S. Mentasti, H. Quan *et al.*, "Advancements in context recognition for edge devices and smart eyewear: Sensors and applications," *IEEE Access*, 2025.

[11] S. C. K. Shek, "Artificial intelligence based image recognition using limited-resource hardware for an aerial drone application," 2025.

[12] A. K. Bhowmik, "Virtual and augmented reality: Human sensory-perceptual requirements and trends for immersive spatial computing experiences," *Journal of the Society for Information Display*, vol. 32, no. 8, pp. 605–646, 2024.

[13] F. Li, Y. Liu, J. Qi, Y. Du, Q. Wang, W. Ma, X. Xu, and Z. Zhang, "Ps5-net: a medical image segmentation network with multiscale resolution," *Journal of Medical Imaging*, vol. 11, no. 1, pp. 014 008–014 008, 2024.

[14] Y. Yang, "Smartphone photography and its socio-economic life in china: An ethnographic analysis," *Global Media and China*, vol. 6, no. 3, pp. 259–280, 2021.

[15] Y. Chuan, "Application of indoor thermal energy cycle and visual communication simulation scene based on machine vision and optical image enhancement," *Thermal Science and Engineering Progress*, vol. 55, p. 102994, 2024.

[16] J. Huang, Y. Xu, Q. Wang, Q. C. Wang, X. Liang, F. Wang, Z. Zhang, W. Wei, B. Zhang, L. Huang *et al.*, "Foundation models and intelligent decision-making: Progress, challenges, and perspectives," *The Innovation*, 2025.

[17] M. Brown, "Color processing for digital cameras," *Fundamentals and applications of colour engineering*, pp. 81–98, 2023.

[18] M. N. Ahimbisibwe and N. Mpekoa, "Mobile phone firmware and hardware hacking detection system," in *International Conference on Cyber Warfare and Security*. Academic Conferences International Limited, 2025, pp. 641–649.

[19] R. Arakawa, K. Maeda, and H. Yakura, "Supporting experts with a multimodal machine-learning-based tool for human behavior analysis of conversational videos," *arXiv preprint arXiv:2402.11145*, 2024.

[20] A. Agarkar, S. Shrivastava, D. Bangde, and D. Bhise, "Cnn based image-denoising techniques-a review," in *2024 3rd International Conference on Sentiment Analysis and Deep Learning (ICSADL)*. IEEE, 2024, pp. 146–152.

[21] R. Wang, Y. Zhang, W. Wan, X. Li, and M. Chen, "Predicting multiscale information diffusion via minimal substitution neural networks," in *IEEE INFOCOM 2024-IEEE Conference on Computer Communications*. IEEE, 2024, pp. 1890–1899.

[22] M. Liu, S. He, S. Lin, and B. Wen, "Dual-head genre-instance transformer network for arbitrary style transfer," in *Proceedings of the 32nd ACM International Conference on Multimedia*, 2024, pp. 6024–6032.

[23] J. Dan, W. Liu, M. Liu, C. Xie, S. Dong, G. Ma, Y. Tan, and J. Xing, "Hogda: Boosting semi-supervised graph domain adaptation via high-order structure-guided adaptive feature alignment," in *Proceedings of the 32nd ACM International Conference on Multimedia*, 2024, pp. 11 109–11 118.

[24] D. Kwon, J. Gwak, and J. Yu, "Long range cctv based context-aware petroglyphs surface change classification using dual attention-guided pyramid siamese network," *IEEE Access*, 2025.

[25] Y. Qi, Q. Zhang, X. Lin, X. Su, J. Zhu, J. Wang, and J. Li, "Seeing beyond noise: Joint graph structure evaluation and denoising for multimodal recommendation," in *Proceedings of the AAAI Conference on Artificial Intelligence*, vol. 39, no. 12, 2025, pp. 12 461–12 469.

[26] Y. Wang, T. Xu, Z. Fan, T. Xue, and J. Gu, "Adaptiveisp: Learning an adaptive image signal processor for object detection," *Advances in Neural Information Processing Systems*, vol. 37, pp. 112 598–112 623, 2024.

[27] M. Yoshimura, J. Otsuka, A. Irie, and T. Ohashi, "Dynamicisp: dynamically controlled image signal processor for image recognition," in *Proceedings of the IEEE/CVF International Conference on Computer Vision*, 2023, pp. 12 866–12 876.

[28] N. Jiang, B. Sheng, P. Li, and T.-Y. Lee, "Photohelper: portrait photographing guidance via deep feature retrieval and fusion," *IEEE Transactions on Multimedia*, vol. 25, pp. 2226–2238, 2022.

[29] S. Moghaddam and M. Ester, "Aqa: aspect-based opinion question answering," in *2011 IEEE 11th International Conference on Data Mining Workshops*. IEEE, 2011, pp. 89–96.

[30] A. Mi, W. Luo, Y. Qiao, and Z. Huo, "Rethinking zero-dce for low-light image enhancement," *Neural Processing Letters*, vol. 56, no. 2, p. 93, 2024.

[31] M. Sandler, A. Howard, M. Zhu, A. Zhmoginov, and L.-C. Chen, "Mobilenetv2: Inverted residuals and linear bottlenecks," in *Proceedings of the IEEE conference on computer vision and pattern recognition*, 2018, pp. 4510–4520.